\documentclass[letterpaper, 10 pt, conference]{ieeeconf}  

\IEEEoverridecommandlockouts

\usepackage{graphics} 
\usepackage{xcolor}
\usepackage{epsfig} 
\usepackage{mathptmx} 
\usepackage{times} 
\usepackage{amsmath} 
\usepackage{amssymb}  
\usepackage[ruled,vlined]{algorithm2e}
\usepackage{svg}
\usepackage{color}
\usepackage{hyperref}
\usepackage{graphicx}
\usepackage{caption}
\usepackage{subcaption}
\newsavebox{\bigleftbox}
\usepackage{CJKutf8}
\usepackage{amsfonts}
\usepackage{setspace,lipsum}
\usepackage{cite}

\newsavebox{\bigimage}

\newtheorem{assumption}{Assumption}
\newtheorem{problem}{Problem statement}

\title{\LARGE \bf{
Teleoperated aerial manipulator and its avatar. Part 1: Communication, system's interconnection, control, and virtual world}}

\author{Rodolfo Verdín, Germán Ramírez, Carlos Rivera, and Gerardo Flores$^{*}$
\thanks{This work was partially supported by CONACYT-FORDECYT under grant 292399.}
\thanks{All the authors are with the Laboratorio de Percepción y Robotica, Center for Research in Optics, Leon, Guanajuato, Mexico, 37150. \newline
        {{$*$} Corresponding author: \tt\small gflores@cio.mx}}
}

\begin{document}

\maketitle
\thispagestyle{empty}
\pagestyle{empty}
\begin{abstract}
The tasks that an aerial manipulator can perform are incredibly diverse. However, nowadays the technology is not completely developed to achieve complex tasks autonomously. 
That’s why we propose a human-in-the-loop system that can control a semi-autonomous aerial manipulator to accomplish these kinds of tasks. Furthermore, motivated by the growing trend of virtual reality systems, together with teleoperation, we develop a system composed of: an aerial manipulator model programmed in PX4 and modeled in Gazebo, a virtual reality immersion with an interactive controller, and the interconnection between the aforementioned systems via the Internet. This research is the first part of a broader project. In this part, we present experiments in the software in the loop simulation. The code of this work is liberated on our GitHub page. Also, a video shows the conducted experiments.
\end{abstract}

\section*{Supplementary material}
The implementation of our system is released on GitHub and is available under the following link: \newline
\url{https://github.com/Rodolfo9706/VR-teleoperated-aerial-manipulator.git} 

In addition, this letter has a supplementary video material available at \newline \url{https://youtu.be/Ur4sNFR9U-Y} , provided by the authors.

\section{Introduction}
Some of the most important problems for aerial manipulators are the control, and the complexity in achieving a variety of grasping and manipulation tasks \cite{Ten-Questions,Recent-Researches,Aerial-manipulation-a-literature-survey}. For that, several efforts have been conducted for robot teleoperation , \cite{ref4}\cite{ref5}. One of the major problems of teleoperated systems is the difficulties encountered by the human pilot while he/she is performing a task due to its visualization restriction \cite{10.1145/3232232}. This is especially true for moving robots \cite{i2.pdf}. In this work, we aim to solve the problem of interaction between the human operator and the scene seeing by an aerial manipulator robot by using a robot’s avatar and a virtual reality world which is a copy of the real one.
%


\begin{figure}
\begin{minipage}[t]{.59\columnwidth}
  \vspace*{\fill}
  \centering
  \includegraphics[width=\columnwidth]{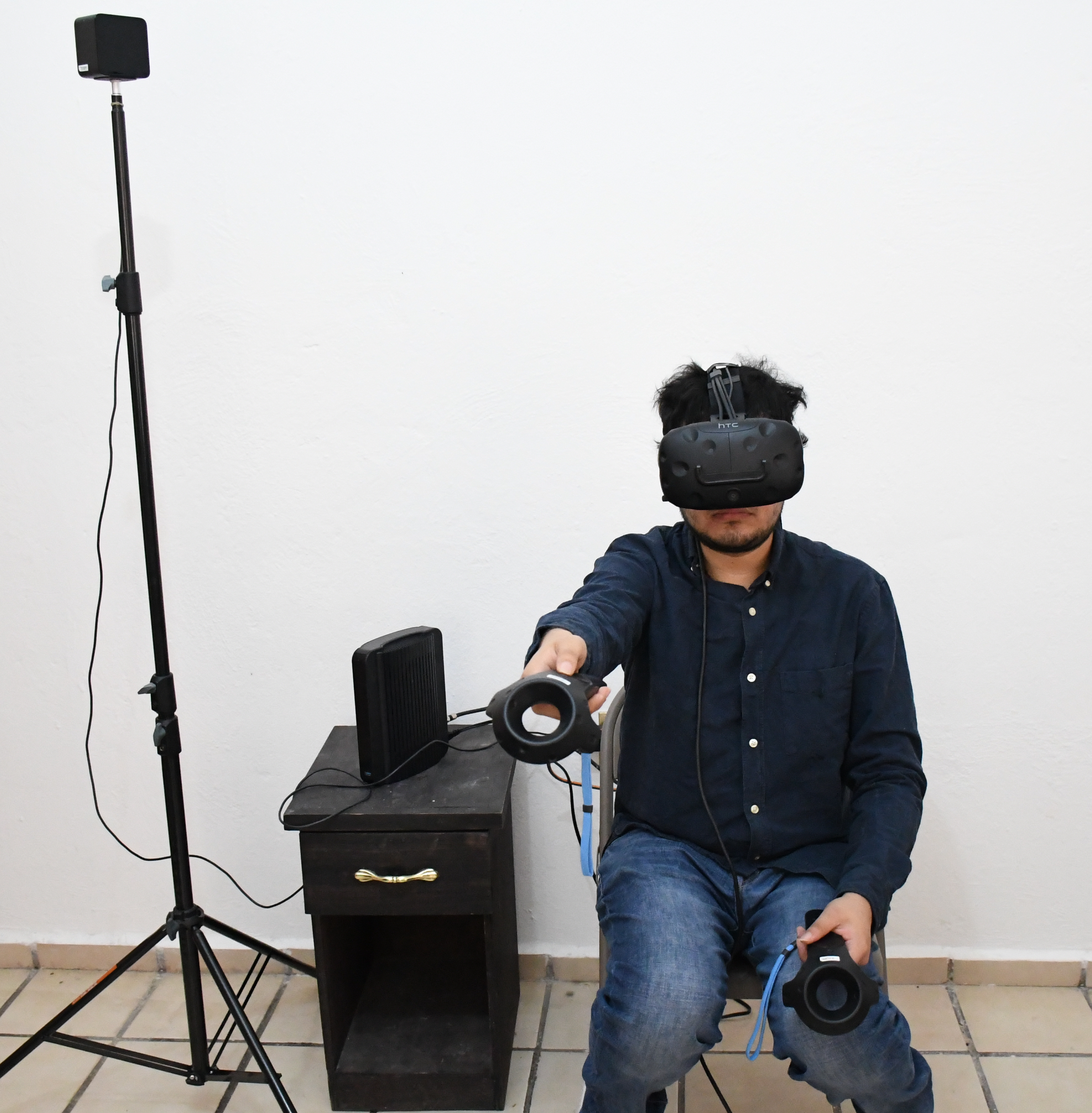}
  \subcaption{Virtual reality immersion.}
  \label{fig:test1}
\end{minipage}%
\hfill
\begin{minipage}[t][\ht\bigleftbox][s]{0.39\columnwidth}
  \vspace*{\fill}
  \centering
  \includegraphics[width=\columnwidth]{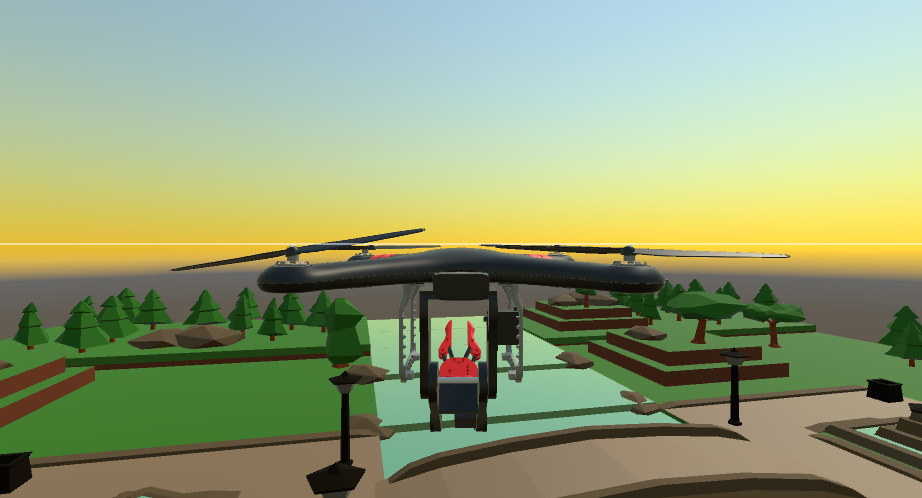}
  \subcaption{Virtual reality environment}
  \label{fig:test2}\par \vspace{5mm}
  \includegraphics[width=\columnwidth]{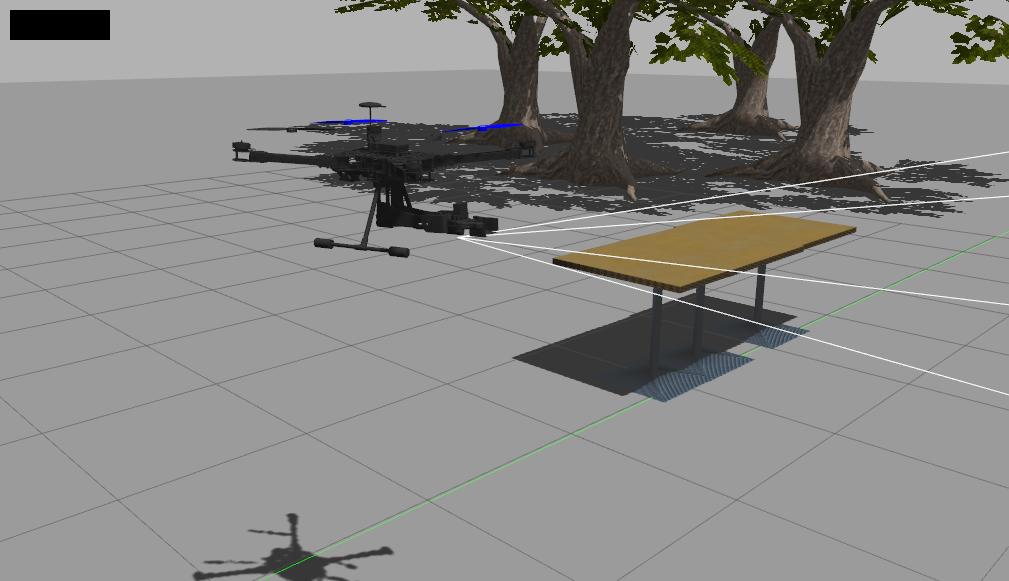}
  \subcaption{3D simulation}
  \label{fig:test3}
\end{minipage}
\caption{This system is composed of: (a) The use of HTC - live to immerse yourself in virtual reality (b) is the view that provides the virtual reality environment made in unity this environment replicates the movements made in (c) which is a simulation of an aerial manipulator performed using px4 autopilot.}
\end{figure}



%
%
\subsection{Problem}
Performing a given task with a teleoperated aerial manipulator seeing the scene through mounted-cameras can be a complicated task for the human operator due to the limitation of vision on the scene. In this approach, the operator can see only what the mounted-camera can see. Using virtual reality is an alternative since you can have a virtual model in a ground control that replicates the movements and orientations of the robot \cite{8981574}. Thus, the operator is able to visualize and control the robot by means of a reconstructed virtual world, knowing at any moment all the scenes of the environment.
\subsection{Contribution}
This paper presents the development and construction of the following system: an aerial manipulator and its avatar which is semi-autonomously commanded via the Internet. The avatar means a system that transfers the robot’s states, actions, presence, and environment to a virtual location in real-time. In this part of the research the contribution is as follows: a) the control algorithm for the robot; b) the software development for the interconnection via the Internet of all the parts of the complete system; c) test in the SITL simulation environment using the PX4 autopilot firmware; d) a demonstration of the experiments in a virtual world created in Unity; e) teleoperated control with a virtual reality headset.
The aforementioned establish the first part of a bigger project. The part II will include the experiments with the real aerial manipulator depicted in Fig. \ref{real-vehicle}; and 2) the SLAM algorithm for reconstructing the environment via SLAM, that we have developed previously. This scene reconstruction will be performed in real-time building the virtual world, which is a copy of the real world seeing by the aerial manipulator. 
%
%
\subsection{Content}
The remainder of this paper is as follows. Section \ref {sec:related} presents the related work of our contribution. In Section \ref{sec:problem} the problem statement is given with some remarks and stating the scope of the article. In addition, we describe the main problems tackled by our approach. Then, in Section \ref{sec:model-control} we develop the modeling and control approach we used for the aerial robot. Section \ref{sec:arch} describes the complete system architecture of our approach including the communication, interconnection, and virtual world. In Section \ref{sec:exp} the experiments that corroborate our approach are presented and explained. Finally, in Section \ref{sec:conclusion} the future work and the conclusions are discussed.
%
\section{Related work}\label{sec:related}
With regard of control, there are several recent efforts in controlling aerial manipulators \cite{[3]}, \cite{[4]}, \cite{[5]}, \cite {9197394}. Some of them considering complex coupled dynamics. We propose a geometric control programmed in the PX4 autopilot and tested in software in the loop simulations (SITL). Such control has demonstrated to be robust enough under the forces and torques exerted by the arm.

With the aim to create a dynamic and intuitive interface between the human and the robot for object manipulation, several applications have been developed in virtual environments. Certain applications include flying UAV systems \cite{r7.pdf} and simulations where the 3D world is reconstructed and gives dimensional feedback while the user is teleoperating \cite{r6.pdf}. In reference \cite{r1.pdf} a monitoring system for old buildings is created through virtual reality through a 3D reconstruction to detect possible structural damage using a UAV. On the other hand in \cite{r2.pdf} vision-based navigation algorithms for UAVs are developed to monitor dependent people through a virtual environment. 

Tasks involving manipulation are presented in \cite{r8.pdf} for maintenance or repairing industrial robots in a VR environment, using an HTC vive device, and \cite{r9.pdf} for complex manipulating tasks using Virtual Reality (VR) sets. For high-risk tasks in \cite{r4.pdf} intuitive and effective control methods based on virtual reality implemented in ROS packages are proposed to teleoperate an underwater robotic arm to manipulate potentially dangerous objects. Regarding aerial manipulation systems win with virtual reality in reference \cite{p2.pdf} a 3D virtual simulator is implemented for collaborative tasks of autonomous and teleoperated navigation in a virtual room. The simulator allows the user to manipulate an object using two robotic arms attached to an aerial vehicle. Haptic and Virtual Reality-Based Shared Control for MAV is presented in \cite{p7.pdf} including an interface that allows safe operation in addition to provide a stable interaction and control between the aerial manipulator and the remote environment. A combined feedback system for an aerial manipulator is presented in \cite{8981574} using VR trackers set in the user's arm and tracking gloves. This framework provides vibrating feedback to control the robotic arm and a Head-Mounted Display to visualize the object. In \cite{10.1145/3232232} the authors propose a new interaction paradigm that provides adaptive views for improving drone teleoperation. This adaptive view provides a user with environment-adaptive viewpoints that are automatically configured to improve safety and provide smooth operation. However, in \cite{10.1145/3232232} the authors only focus their efforts on 3D reconstruction and virtual navigation with the human, not considering an aerial manipulator neither an avatar of it. Unlike these works, our contribution offers a solution for teleoperating systems in which an avatar recreates the movement of the real robot while a human operator controls the real robot. The interconnection of the robot, the human interface, and the avatar is via the Internet. In addition, this work was built using open source code free for the community.

\section{Problem setting}\label{sec:problem}
The problem statement is as follows:
\begin{problem}
\textit{The problem can be devised in two layers. The first layer is that of proposing a way to perform complex tasks in a human-assisted aerial manipulator. The second layer is finding a solution for an enhanced visualization of the environment seeing by the robot and transmitted to the human-operator. In addition, the aerial manipulator must be stable during all the tasks performed by the human operator.}
\end{problem}

For solving this, we propose developing an avatar of a real aerial manipulator endowed with sensors and cameras. Such an avatar receives all the aerial robot’s states, together with the information captured by the embedded cameras. With this, a virtual environment is constructed in real-time identical to that seeing by the aerial manipulator.

The goal of the avatar and constructing a virtual world being a copy of the real world is twofold: 1) for repetitive tasks, one obtains a copy of the real scenario in which several human pilots can be trained to perform desired tasks, and at the same time, recovering all the data [robot states and environment] for eventually training an artificial-intelligence-based control; and 2) with the reconstructed virtual world, the robot can easily navigate and perform tasks that in the case of not having this virtual world could be impossible. If one thinks of drones with mounted cameras commanded by First Person View with a human pilot, one notices that these systems could be an option for our approach. However, those systems cannot reconstruct a copy of the navigated environment and as a consequence they lack of a virtual world. Using FPV systems can cause loose of orientation in the pilot, or even damage in the robot. We are trying to avoid such possible catastrophes with our proposal.

Motivated by the growing trend of virtual reality systems together with teleoperation, we develop a system composed of:
\begin{itemize}
\item[1.]The aerial manipulator. In this part of the research we construct a Gazebo model connected and controlled to PX4 firmware.
\item[2.]Control algorithm to stabilize the UAV during all the scenarios. This is a geometric control programmed in the PX4 firmware.
\item[3.]A virtual world constructed in Unity that simulates the copy of the real world represented in Gazebo. The virtual world includes a virtual copy of the aerial manipulator.
\item[4.]Communication system. This is responsible to send all the control commands from the HTC vive to the aerial manipulator via Internet. Then, this system also sends such signals to the avatar, which copies the behavior of the aerial manipulator presented in Gazebo.
\end{itemize}
%
\begin{figure}
    \centering
    \includegraphics[width=\columnwidth]{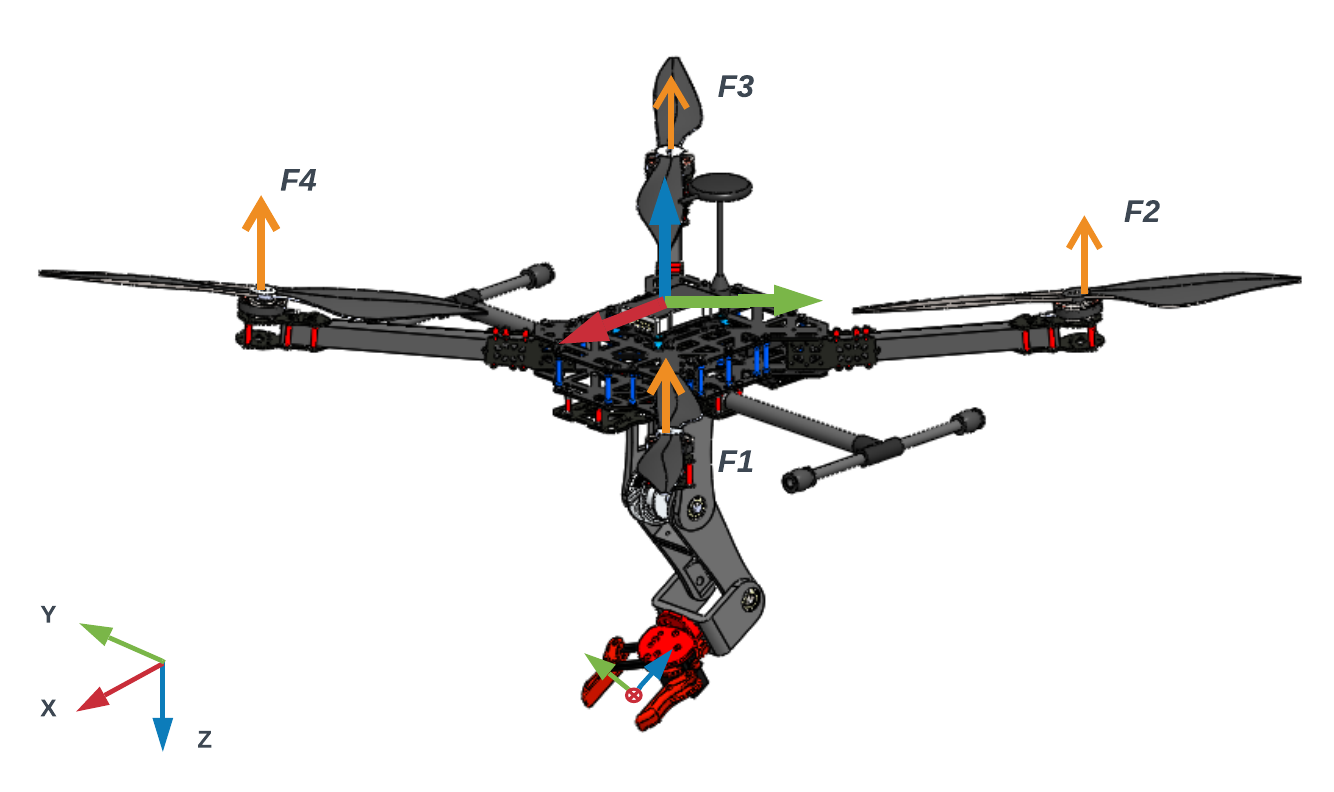}
    \caption{Aerial manipulator CAD model.}
    \label{fig:CAD model}
\end{figure}
%
\section{Modeling and control}\label{sec:model-control}
\subsection{Modeling}
We model the aerial manipulator considering the following assusmptions:
\begin{assumption}\label{ass:moments}
The movement of the manipulator produces unknown moments in the quadrotor frame.
\end{assumption}
\begin{assumption}\label{ass:forces}
The quadrotor serves as a moving platform for the position of the arm. Thus, the arm position is partially controlled by the quadrotor's pose, in this way the forces exerted by the quadrotor frame to the arm are always controlled.
\end{assumption}
\begin{assumption}\label{ass:arm-control}
The arm is totally controlled by the operator, only a simple PD control is implemented in each of the arm joints.
\end{assumption}
Thus, the aerial manipulator is modeled as follows
\begin{eqnarray}
\dot{x} &=& v \\
\dot{v} &=& g {e}_3 - \frac{f}{m}R {e}_3 +  F_{a}(t) \label{eq:pos_dyn}  \\
\dot{R} &=& R\hat{\Omega}  \label{eq:att1} \\
\dot{\Omega} &=&   -J^{-1}\Omega \times J\Omega + J^{-1}\tau + T(t) \label{eq:att2}
\end{eqnarray}
where $\hat{(\cdot)} : \mathbb{R}^3 \rightarrow \mathfrak{so}(3)$ is $\hat{x} = \begin{pmatrix}
0 & - x_3 & x_2 \\ x_3 & 0 & -x_1 \\ -x_2 & x_1 & 0
\end{pmatrix} $ with $x = [x_1 , \ x_2 , \ x_3 ]^{\intercal}$ in which $\mathfrak{so}(3)$ is a Lie algebra in $SO(3)$. Such operation has an inverse given by $\lor : \mathfrak{so}(3) \rightarrow \mathbb{R}^{3} $ which basically transforms a skew-symetric matrix into a vector in $\mathbb{R}^{3}$. $(F_a,T_a)$ are the unknown forces and moment vectors exerted by the arm to the quadrotor frame. The rotation matrix $R$ is given by \cite{Etkin}
\begin{equation}
R = \begin{pmatrix}
c_{\theta}c_{\psi} & s_{\phi}s_{\theta}c_{\psi} - c_{\phi}s_{\psi} & c_{\phi}s_{\theta}c_{\psi} + s_{\phi}s_{\psi}\\
c_{\theta}s_{\psi} & s_{\phi}s_{\theta}s_{\psi} + c_{\phi}c_{\psi} & c_{\phi}s_{\theta}s_{\psi} - s_{\phi}c_{\psi}  \\
- s_{\theta} & s_{\phi}c_{\theta} & c_{\phi}c_{\theta}
\end{pmatrix} 
\end{equation}

\subsection{Control}
Let define
\begin{equation}\label{eq:poserror}
e_p = x-x_d, \ e_v = \upsilon - \upsilon_d
\end{equation}
then we implemente the control
\begin{equation}\label{eq:f}
f= m\|g\hat{e}_3 + K_ve_v + K_p e_p - \ddot{x}_d \|
\end{equation}

The attitude control is given by
\begin{equation}\label{eq:tau}
    \tau = - k_R e_R - K_{\Omega}e_{\Omega}
\end{equation}
with 
\begin{equation}
e_{R} = \frac{1}{2} \left( R_d^{\intercal} R - R^{\intercal} R_d  \right)^\lor \in \mathbb{R}^{3} , \ e_{\Omega} = \Omega - R^{\intercal}R_d \Omega_d \in \mathbb{R}^{3}.
\end{equation}
\section{System architecture}\label{sec:arch}
\begin{figure}
    \centering
    \includegraphics[width=\columnwidth]{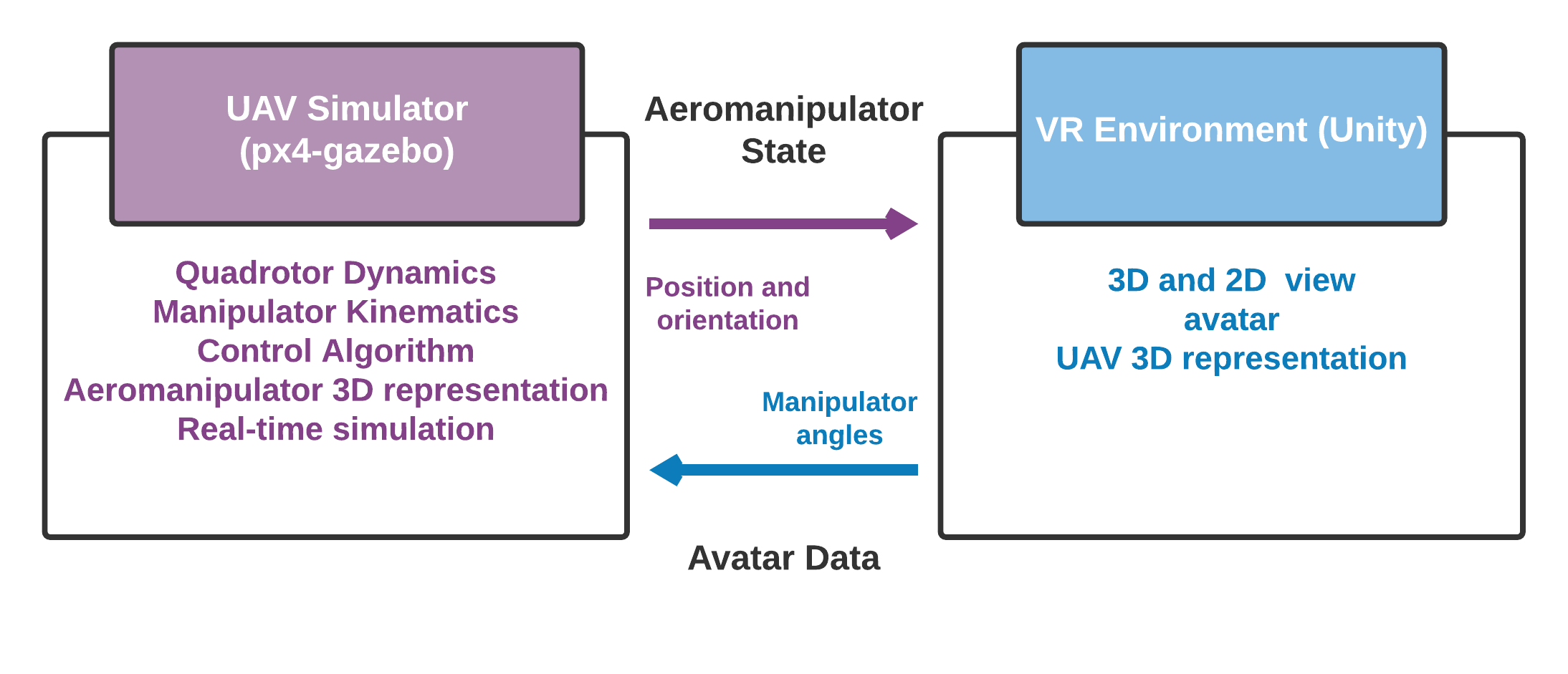}
    \caption{Diagram of the aerial manipulator system visualized in a virtual environment.}
    \label{fig:graph}
\end{figure}

The first element that needs to be added in both virtual environments (Gazebo, Unity3D) is the aerial manipulator 3D CAD model. Gazebo works on Ubuntu 18.04 and Unity works on Windows 10 OS and both of the environments communicate through common MAVROS messages which are included inside their respective topics. The topics to be employeed are LocalPosition and MountControl, inside the LocalPosition topic the element "pose" collects the local quadrotor positions through GPS and this information is sent through a string type message. The MountControl topic is used to publish and subscribe to the robotic arm's orientation or any other actuator that can be included in the vehicle model. Attitude information is published from Gazebo and the PX4 works as a subscriber to display that information in the Gazebo simulation. The visual system structure is described in Fig. \ref{fig:graph}. The system architecture is divided in three subsections: A) Unity; B) Gazebo \& PX4; and C) communication (mavros).
%
%
%
\subsection{Unity} 
\begin{figure}
    \centering
    \includegraphics[width=\columnwidth]{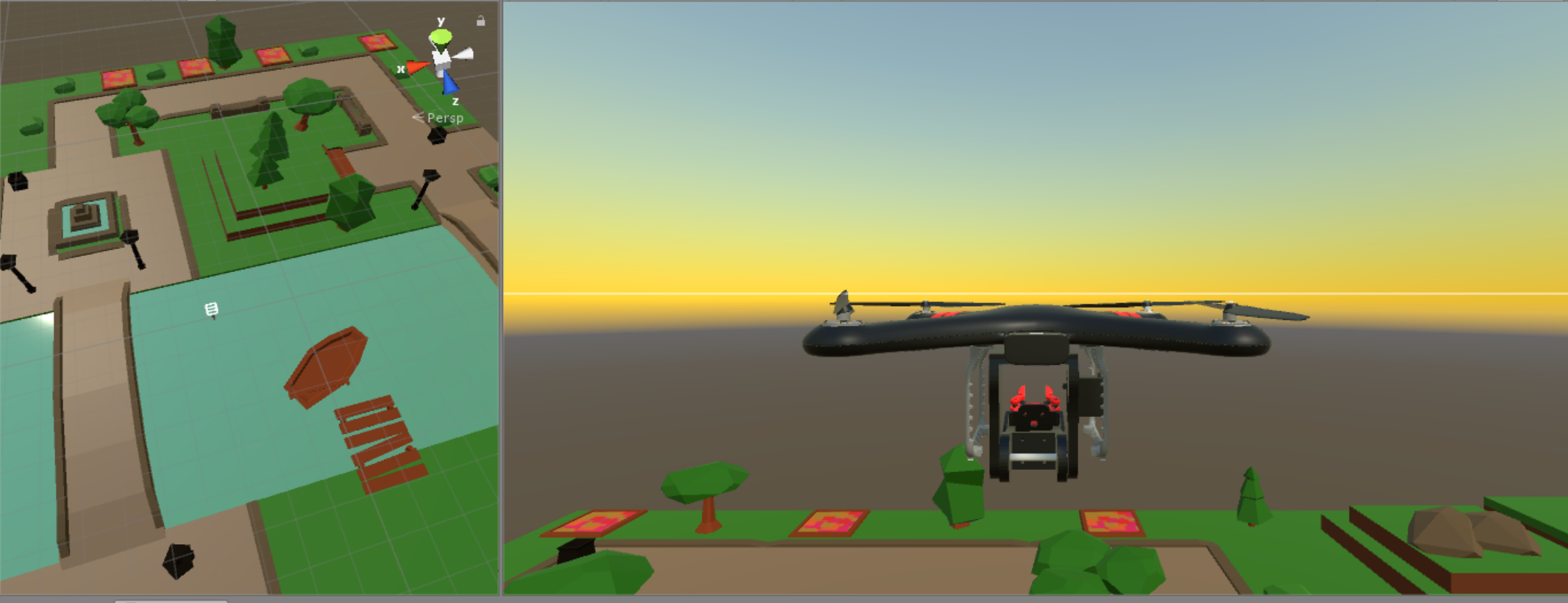}
    \caption{The avatar and the virtual world made in Unity.}
    \label{VR viewer}
\end{figure}

The purpose of designing this environment in Unity is to create a special remote teleoperation ground station. It allows the use of virtual reality devices to visualize the environment and get a robot's position feedback in the real world. The aerial manipulator model in Unity contains several apps. These apps work individually and interact with each other to create and send the robot's states depending on the input and the force exerted over the aerial manipulator model. The objects inside Unity are divided by hierarchy keeping the reference over the same objects of the model to place them or create the necessary force to move them in a local or global form as it is required.  Therefore, the main reference of the model is placed on the UAV body followed by each link of the robotic arm and each revolute joint representing every degree of freedom that moves individually. The main reference of the model contains one app to get the position of the vehicle and the other app is to communicate Unity and Gazebo. Depending on the position input the first code generates the necessary force to be applied on each rotor to get the desired attitude ($\phi$,$\theta$,$\psi$) leading to the desired position obtained from the position data in the Gazebo simulation. This virtual environment in Unity3D is a recreation of the real world where the aerial manipulator is moving involving the information concerning building dimensions and obstacles that might need to be avoided in the real world. In Fig. \ref{VR viewer} it is shown the virtual reality world and its two-screen views. These views aim to facilitate the tasks performed by the robot through the operator and represent the avatar's view.

%
\subsection{The Gazebo model}
Gazebo is an open-source 3D robotics simulator that provides elements such as sensors, actuators control, cameras, simulation tools, and realistic dynamics of each model. The purpose to make this virtual model is to work on a SITL simulation (software in the loop) that allows testing and simulating the aerial manipulator before being tested in the real world (which is the subject of the part 2 of this project). The Gazebo model of the aerial manipulator is programmed by us with the PX4 firmware, which is one of the most used autopilots by the robotics community. PX4 firmware contains packages that integrate gazebo to perform SITL that facilitates the design of the vehicle and the control implementation. To communicate with some external software/hardware, a communication protocol capable of sending and receiving messages between different topics is needed. Therefore, in this case MAVROS (mavlink / ROS) was the best choice to create Python or C++ scripts to extract or send information through the PX4 - Gazebo network we have created. In parallel, the ground station Qgroundcontrol is employed to work as a command center, where technical information about the aerial vehicle is displayed.

\begin{figure}[t]
    \centering
    \includegraphics[width=\columnwidth]{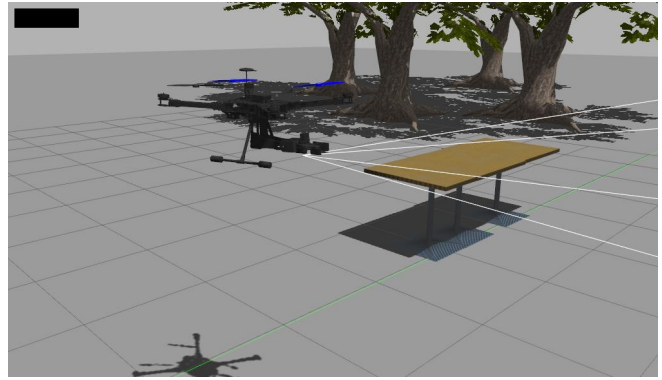}
    \caption{Model simulation of the aerial manipulator in Gazebo using PX4 autopilot.}
    \label{px4-model}
\end{figure}

We construct the aerial manipulator in Gazebo based on an existing model called Typhoon ($typhoon\_h480$ in the PX4 firmware). The Typhoon model is a Hexa-rotor with an embedded gimbal. The model \textit {$ typhoon\_h480 $} is in the PX4 firmware directory \textit{$ src / tools/sitl\_gazebo/models/typhoon\_h480 $} where a document meets SDF format. This contains all the aerial vehicle's features such as color, collisions, visual, sensors, actuators, etc. In the same Firmware, a folder called Meshes contains the frame, the body link, and impellers in SITL format. To transform into a different vehicle, it is necessary to add the Collada files (*.dae) to the meshes folder. For the dimensions in the propellers and the manipulator, the collisions were modified. The created model is depicted in Fig. \ref{px4-model}.
%

All the necessary commands and prompts to run and install the environment can be found in our GitHub page referenced at the beginning of the document.

\subsection{Mavros \& comunication}
MAVROS is a ROS node that allows communication through mavlink protocols containing several topics, each of them contains specific information about sensors and actuators from the aerial robot. As was previously mentioned, the required topics for the aerial manipulator are the \textit{LocalPosition} and \textit{MountControl} for the position output (subscriber), and the robotic arm position inputs (publisher), respectively. 
 \begin{figure*}[h]
    \centering
    \includegraphics[width=\textwidth]{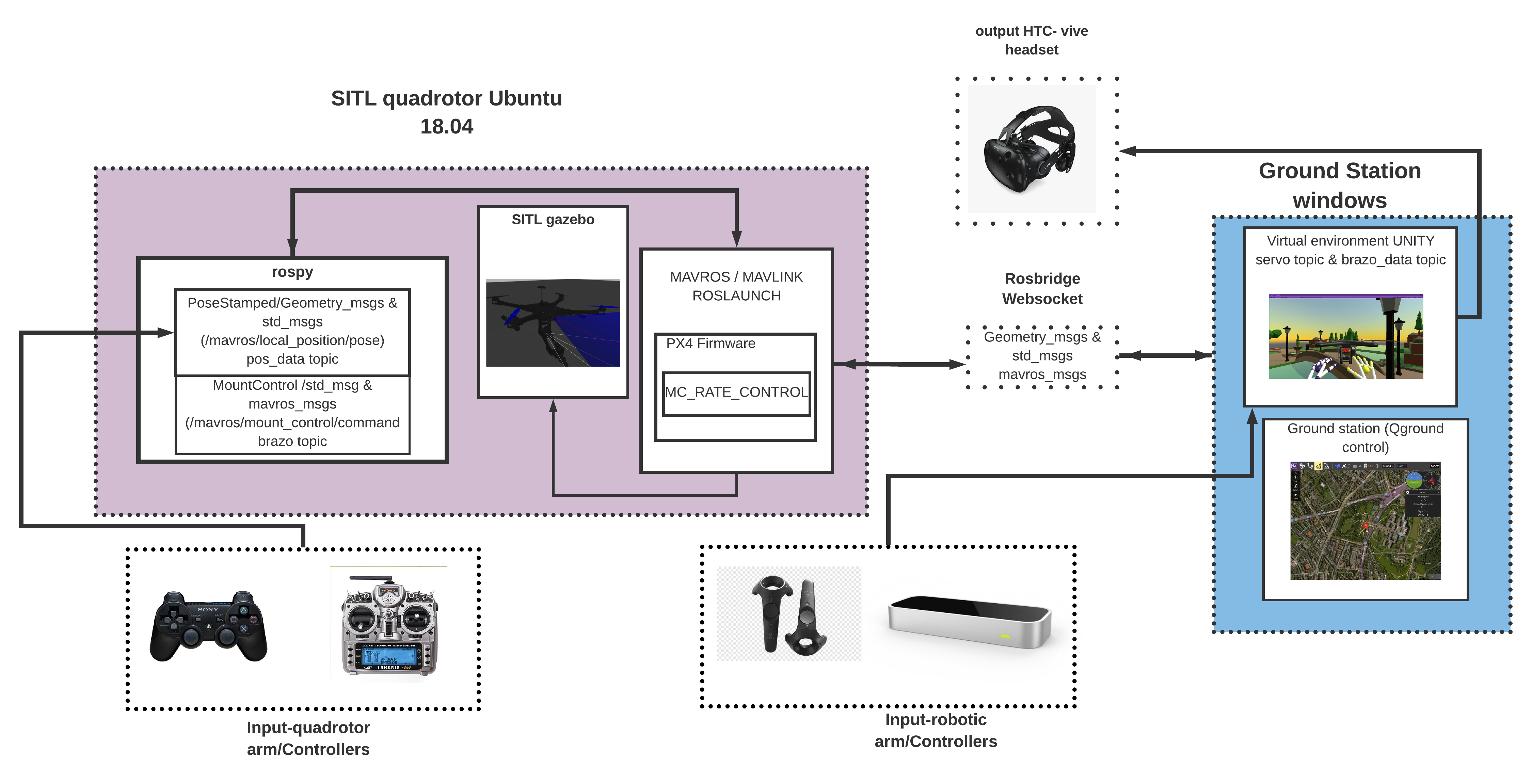}
    \caption{The color purple represents the gazebo model of the aerial manipulator with the ROS topics and nodes used from PX4. In the center' of the figure is the interconnection using rosbridge. In color blue are the typical apps used in Unity for the visualizer in the virtual reality world.}
    \label{fig:Diagram}
\end{figure*}

To get data from \textit{LocalPosition} a python script was created using rospy. It allows linking python with ROS. Then, the geometry message was defined to establish communication with the aerial robot and be published in the \textit{servo} message. To allow Unity to get the message, it is used a ROSBridge protocol, which is a WebSocket. Then, it subscribes to \textit{servo} message to get the positions messages via the Internet. To get and read the message from Unity, a script is created to communicate with Ubuntu using an IP address, and then receiving the data subscribing to the \textit{servo} message. To do this process, JSON-formatted codes were used, Once Unity gets the positions, it moves the vehicle in the virtual environment to get to the same position from the message.

To get the positions from the robotic arm, a similar process is made. The topic used in this case is \textit{MountControl}. Another python script is created to publish 3 variables for each joint of the robotic arm. Gazebo reads those 3 variables and moves the robotic arm to the indicated position given by the human operator. Then, Unity subscribes to a message called \textit{data}, created inside the same script, to get the robotic arm information to move the manipulator in the virtual environment. A complete diagram of the communication structure is shown in Fig. \ref{fig:Diagram}.




\section{Experimental results}\label{sec:exp}
This section presents experimental results of the approach described in the last sections. The experiments consist in teleoperating an aerial manipulator trough the HTC vive VR headset via the Internet connection. Furthermore, we show the avatar of the aerial robot mimics the dynamic behavior of the aerial manipulator. Also, we test the control law given by \eqref{eq:f} and \eqref{eq:tau}.

The computer used for Gazebo simulation is an Intel Core i7-7820HK laptop with 32 Gb RAM and a GPU GeForceGTX 1070. For the virtual reality environment, we used an AMD A12-97209 laptop with 12 Gb RAM at 2.70 GHz, with no GPU.

\subsection{Control algorithm}
\begin{figure}[t]
    \centering
    \includegraphics[width=\columnwidth]{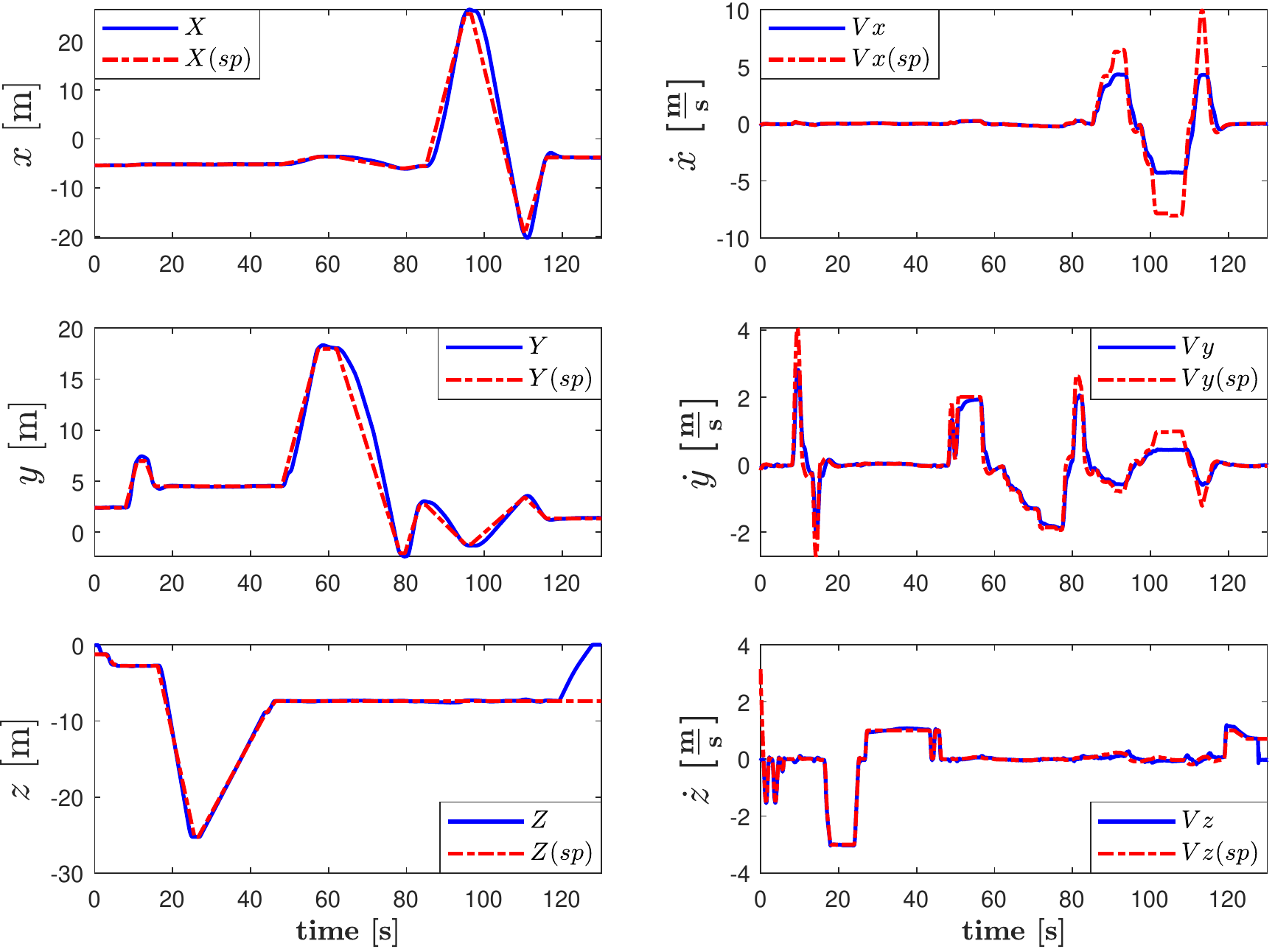}
    \caption{Robot's position and linear velocity during software in the loop simulation conducted in Gazebo under control \eqref{eq:f}. The control law is programmed in PX4 firmware.}
    \label{graph1}
\end{figure}

\begin{figure}[t]
    \centering
    \includegraphics[width=\columnwidth]{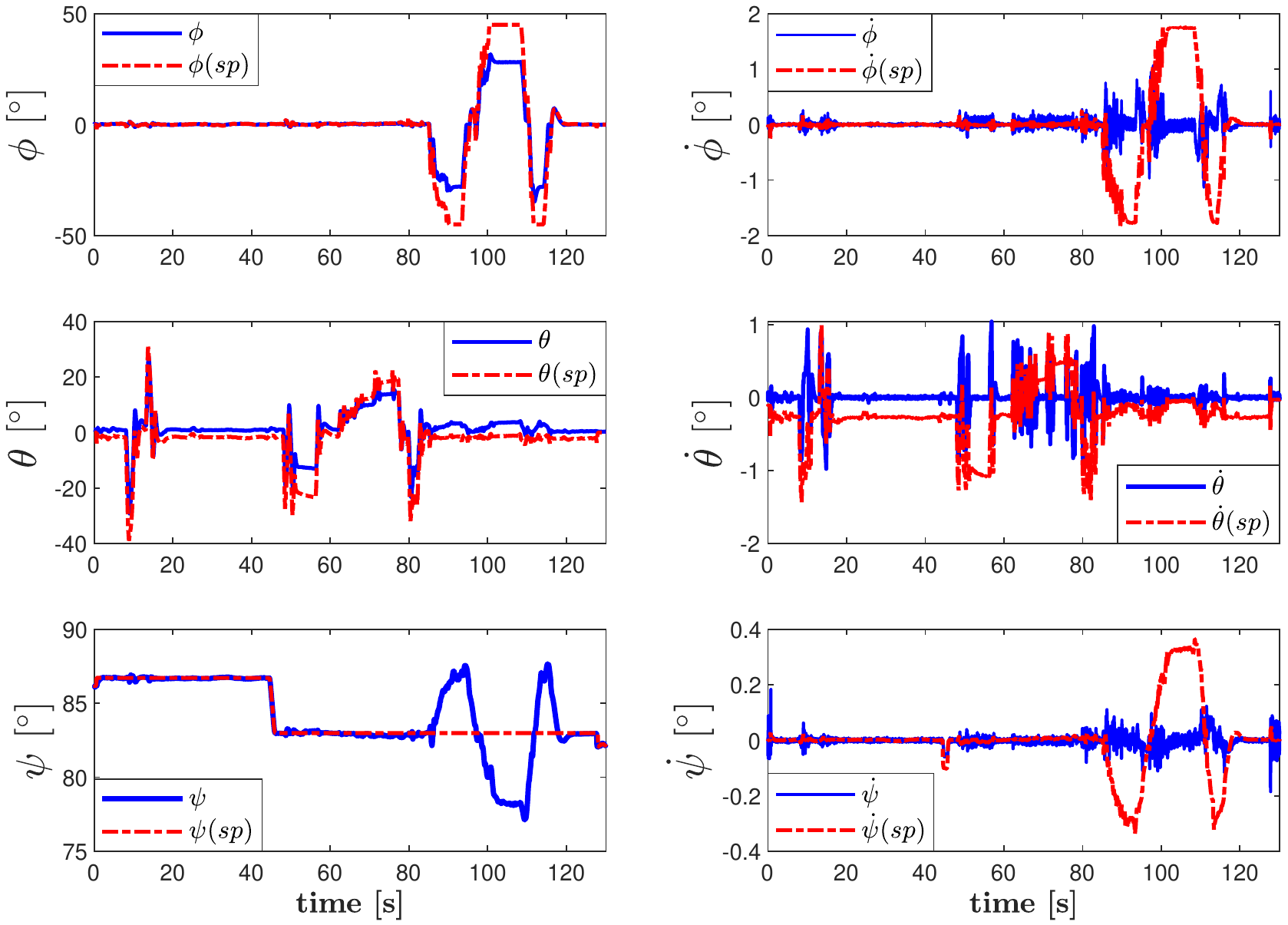}
    \caption{Aerial manipulator's attitude and angular velocities during software in the loop simulation conducted in Gazebo under control \eqref{eq:tau}. The control law is programmed in PX4 firmware.}
    \label{graph2}
\end{figure}

The objective of the first experiment is to measure the error between the setpoint position and the actual position of the vehicle, under the action of controls \eqref{eq:f} and \eqref{eq:tau}. We test the control response under the following teleoperated movements: left, right, up, down and several rotations. While we move the manipulator in several directions. The results are as depicted in Fig. \ref{graph1} where robot’s pose and velocities are given. 
%
Angular positions and angular velocities are shown in Fig. \ref{graph2}. Notice how the attitude is 
%
\subsection{Response's time}
One of the problems in teleoperated control is maintaining a response time with almost zero delays. In this part, we report the results in this regard. The desired behavior is that of maintaining an instantaneous response in the avatar dynamics, each time the aerial manipulator moves. For that, the network performance and the interconnection system must be stable. In Fig. \ref{graph2} it is plotted the robot's position and the avatar position during flight experiments. According to the results, there is a good fidelity in the robot movements, while the vehicle delay is around 0.5 seconds. Notably, there are times when there is a greater delay. To solve this, one can use a computer with better specs.
\begin{figure}[t]
    \centering
    \includegraphics[width=\columnwidth]{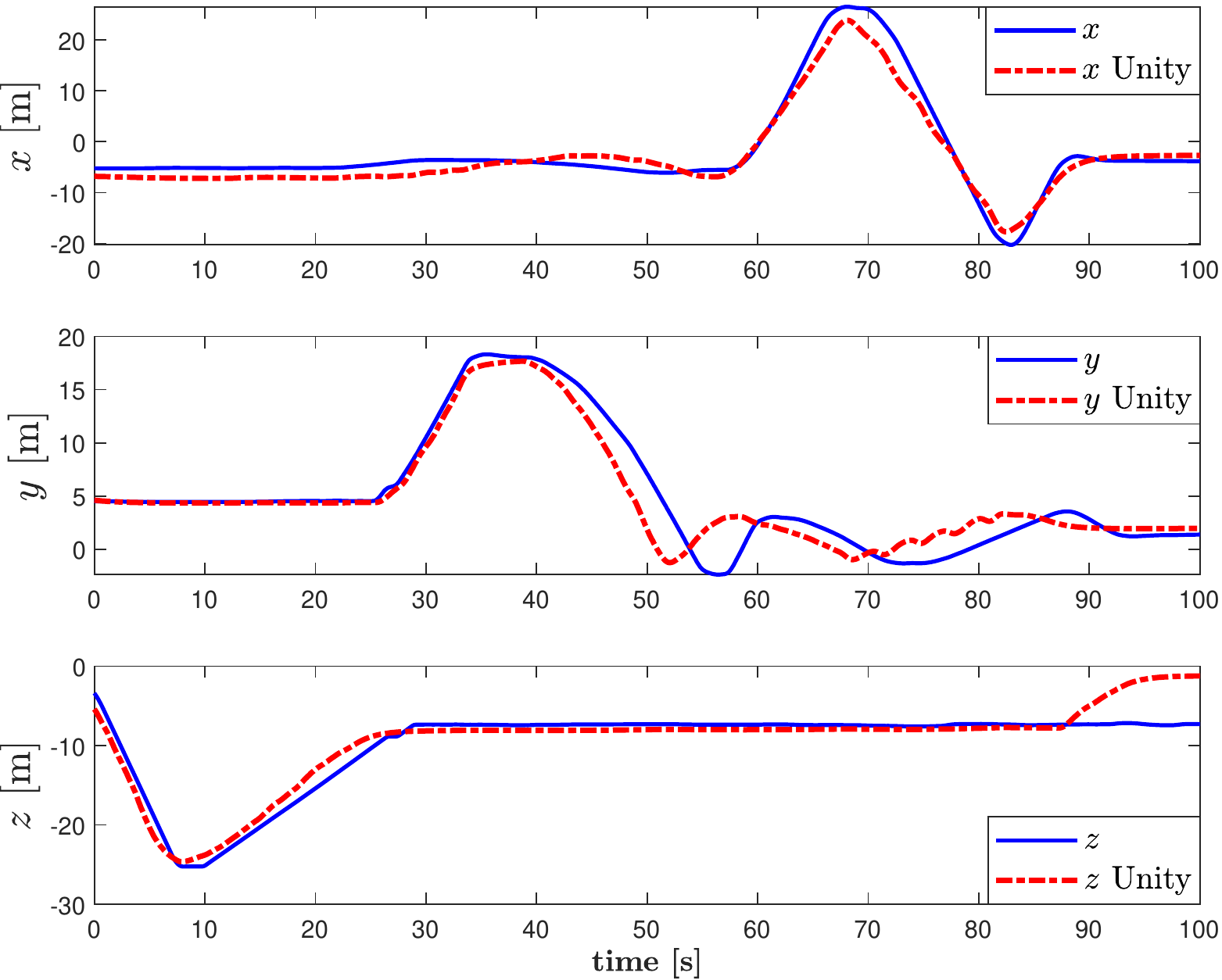}
    \caption{Blue color positions in the vehicle and red gazebo positions of the avatar in Unity}
    \label{graph3}
\end{figure}
%
\subsection{Pick and place experiment}
In this experiment, the aerial robot is teleoperated through the HTC vive. The task to perform is that of taking an object and transporting it to a given point chosen by the operator. This is performed in the Gazebo environment while running the control algorithm. The avatar copies the task and the environment. The results demonstrate the control performance and robustness to a mass variation and under forces and moments generated by the arm and exerted to the drone. The mass of the object is 160 grams.

A video of the experiments is available in the following link \newline
\url{https://youtu.be/Ur4sNFR9U-Y}

\begin{figure}[h]
    \centering
    \includegraphics[width= 9 cm]{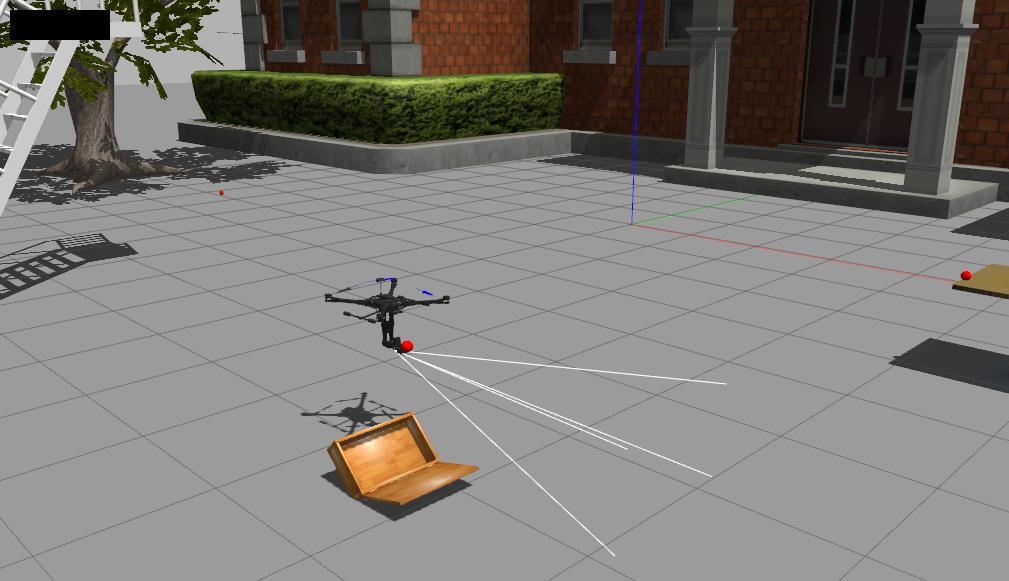}
    \caption{Pick and place experiment}
    \label{pick}
\end{figure}
%
\section{Conclusions}\label{sec:conclusion}
A teleoperated aerial manipulator was simulated in a virtual environment in Unity 3D and Gazebo. The simulation contains the vehicle dynamics and the kinematics of the manipulator, as well as the control programmed in the PX4 firmware in Gazebo. Also, it is teleoperated remotely by commands transmitted via ROSBridge protocol (WebSockets). This allowed the VR application to visualize in real-time the states of the aerial manipulator. The time response can be improved using high-performance computer equipment. 

\subsection{Future work}
\begin{figure}[t]
    \centering
    \includegraphics[width=\columnwidth]{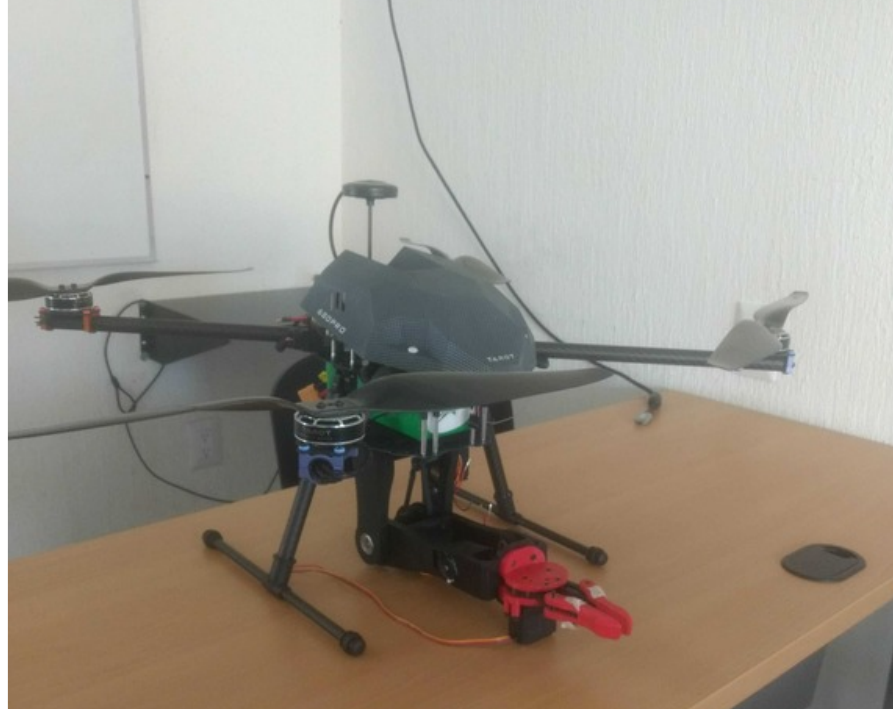}
    \caption{Aerial manipulator protoype constructed in the LAB. This robot will be used in the second part of this research.}
    \label{real-vehicle}
\end{figure}

For the second part of this research, the experiments will be conducted in the real aerial manipulator developed at the lab and depicted in Fig. \ref{real-vehicle}. In that work, a SLAM system will be implemented to reconstruct the virtual environment with real dimensional and imaging data. Also, a more intuitive sensorial virtual reality system to control the aerial manipulator will be included.
\section*{Acknowledgments}
%
The authors greatly appreciate the comments and time taken by the editor and the anonymous reviewers in evaluating this paper.

\bibliographystyle{ieeetr}
\bibliography{root.bib}
\end{document}